\documentclass[conference]{IEEEtran}

\IEEEoverridecommandlockouts

\usepackage{cite}
\ifCLASSINFOpdf
   \usepackage[pdftex]{graphicx}
   \graphicspath{{./Images/}}
\else

\fi
\usepackage{color}
\usepackage{amsmath}
\usepackage{amsfonts}
\begin{document}
\title{Learning Robust Representations for \\ Automatic Target Recognition}

\author{
  \IEEEauthorblockN{Justin A. Goodwin, Olivia M. Brown, Taylor W. Killian, and Sung-Hyun Son}\thanks{This material is based upon work supported by the Assistant Secretary of Defense for Research and Engineering under Air Force Contract No. FA8702-15-D-0001. Any opinions, findings, conclusions or recommendations expressed in this material are those of the author(s) and do not necessarily reflect the views of the Assistant Secretary of Defense for Research and Engineering.}
  \IEEEauthorblockA{  MIT Lincoln Laboratory \\
  244 Wood St. \\
  Lexington, MA 02420 \\
  \{jgoodwin, olivia.brown, taylor.killian, sson\}@ll.mit.edu \\
}}

\maketitle

\begin{abstract}
  Radio frequency (RF) sensors are used alongside other sensing modalities to provide rich representations of the world. Given the high variability of complex-valued target responses, RF systems are susceptible to attacks masking true target characteristics from accurate identification. In this work, we evaluate different techniques for building robust classification architectures exploiting learned physical structure in received synthetic aperture radar signals of simulated 3D targets.
\end{abstract}

\IEEEpeerreviewmaketitle

\section{Introduction}
\label{sec:intro}

  Active sensors (i.e., radar) can provide autonomous systems with a rich representation of the physical world, which can be used to augment the information collected from traditional static sensors (i.e., cameras). As a radio frequency sensor, radar offers unique capabilities to accurately measure physical attributes that other sensors cannot, such as range to target, radial velocity, and other physical characteristics~\cite{skolnik2008introduction}.  Radar can be used to help with scene characterization and automatic target recognition (ATR) to classify different detected targets (e.g., cars, pedestrians, obstacles) in the presence of different types of clutter (buildings, trees, other noise sources).

  ATR using Synthetic Aperture Radar (SAR) is a common radar application for classifying targets using a sensor mounted on moving vehicles such as aircraft and automobiles. ATR has long been performed with handcrafted features~\cite{yang2005automatic}, an approach that has begun to give way to data-driven approaches following the success of deep learning architectures in image classification~\cite{wilmanski2016modern,profeta2016convolutional,mason2017deep}. However, recent applications of these techniques to radar problems do not explicitly account for the rich physical properties of the signals provided for classification~\cite{guo2017synthetic,morgan2015deep,wagner2017deep}. Using simulated radio wave interactions with 3D targets, we train a model that approximates the relationship between the radar signal response and underlying target class.

  Machine learning models have been shown to be vulnerable to adversarial attacks in which inputs to the model are purposely manipulated in order to produce erroneous results~\cite{biggio2013evasion}. In response, numerous methods have been proposed for generating attacks, building defenses, and measuring the robustness of algorithms to adversarial perturbations~\cite{goodfellow2014explaining,moosavi2016deepfool,wang2016theoretical,xiao2018generating}. SAR systems may also be susceptible to attack given the high variability of possible complex-valued signatures for a given target. This variability results from a number of factors, including diverse environments, sensor parameters, viewing geometries, clutter, target shapes and materials, all of which impact the signal returned to the radar. This variability is difficult to model, and hence difficult to incorporate into the training data for a given radar system. Further, this high variability of possible target signatures leaves opportunity for radar systems to be fooled by an adversary.

  In this work, we evaluate a suite of techniques for building ATR architectures that intend to be robust to adversarial attacks.  The techniques we consider include conditional training based on target pose estimation, feature similarity embedding, and adversarial learning by perturbing the complex-valued target response before processing the image. We evaluate these techniques using physics-based simulations of SAR images for a target shape classification problem, and demonstrate their ability to increase the robustness (and accuracy) of our radar classifier.

  \section{Preliminaries}

\subsection{Neural Networks}

Neural networks, most notably those used in applications of Deep Learning, have emerged as effective near-universal approximators of complex systems and functions. The primary unit of a neural network is called a \emph{neuron}. A neuron, or \emph{node}, can be described as a non-linear mapping over a weighted linear combination of other nodes or some external input. As such, the output of a collection of nodes, described as a \emph{layer} in the network, can be represented as
\begin{equation}
    \tilde{x}^{(k+1)} = \sigma \left(\tilde{W}^{(k)}\cdot\tilde{x}^{(k)} + w_b^{(k)}\right)
\end{equation}
where $\tilde{x}^{(k)}\in\mathbb{R}^n$ is the input vector, $\tilde{W}^{(k)}\in\mathbb{R}^{m\times n}$ is the weight matrix and $w_b^{(k)}$ is the weight bias of the $k^{th}$ layer of the network. $\sigma(\cdot)$ is an element-wise nonlinear mapping, often called an \emph{activation function} that controls for translational invariance as well as introducing the capability of modeling more complex behavior. The output vector $\tilde{x}^{(k+1)}\in\mathbb{R}^n$ is then used as input to the next layer in the network.  The collection of all weight matrices and biases across the layers of a network are known as the \emph{parameters} of the network and may be represented by the variable $\Theta$.

For notation purposes, we subsume the biases $w_b^{(\cdot)}$ into the weight matrices and augment the input $x$ by adding a $1$ as an additional dimension. Thus the parameters of the layer $k$ can be compactly represented by the matrix $W^{(k)}$. The final output of an $N$-layer neural network can be formulated as
{\small
\begin{equation}
    \hat{y} = f_{\Theta}(x) = \sigma\left(W^{(N)} \cdot \sigma\left(W^{(N-1)} \cdots \sigma\left(W^{(0)}\cdot x\right)\right)\right)
\end{equation}
}

The network parameters $\Theta$ are ``learned" through optimization to approximate an unknown function $f:\mathbb{R}^{n_I} \to \mathbb{R}^{n_O}$. This is done by a process called \emph{training} in which the calculated state of the output $\hat{y}$ is compared to target values, $y$, corresponding to the datum as input to the network, $x$. The total deviation from the target value is termed as the \emph{error} of the neural network, the signal of which is then used to update $\Theta$. The standard algorithm for propagating the error through the network is known as \emph{backpropagation}~\cite{rumelhart1985learning}.

Based on the application there are various approaches to measure this error, known as the \emph{loss function}. In practice, the loss function and hyperparameters used by the backpropagation algorithm are the most important aspects, followed by the structure of the nodes and edges that make up the computation graph, of training an accurate neural network and learning a good representation of the collected data. Depending on the complexity of the defined neural network, difficultly of learning the representation of the data, the size of the data set used to train the neural network, etc. influences the number of iterations needed for the backpropagation algorithm to optimize the parameters $\Theta$ appropriately.

\subsection{Robustness and Adversarial Perturbations}

    The goal of any machine learning task is to train the algorithm to perform well on data in the training set, while maintaining performance on data it has not seen before (i.e., maintain high in-sample and out-of-sample accuracy).  This property is referred to as generalization~\cite{abu2012learning}. Out-of-sample performance can be estimated by using a separate test set that is withheld from the training process, or through performing cross-validation of the training set.

\begin{figure}[ht]
        \centering
        \includegraphics[width=0.5\textwidth]{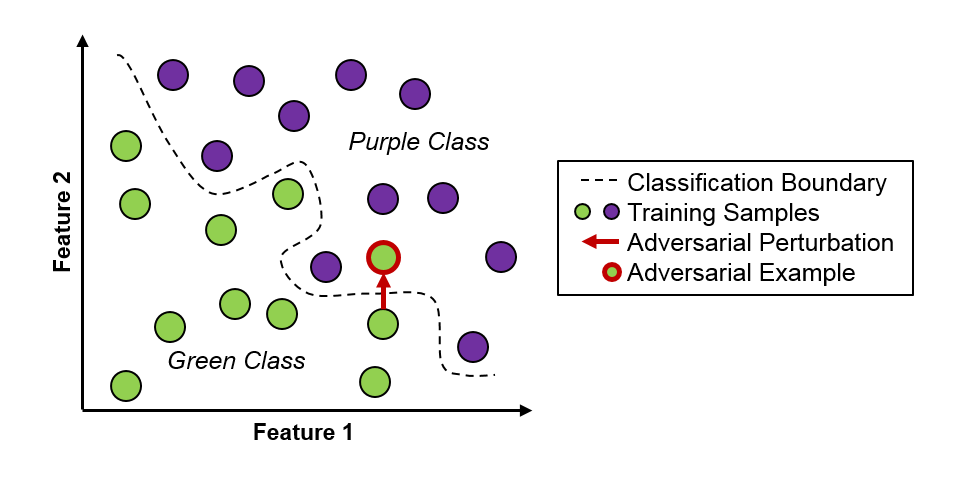}
        \caption{Notional classifier with adversarial perturbation (red arrow) applied to one of the green samples.}
        \label{fig:perturb}
    \end{figure}

    A second property that is desired of a machine learning algorithm is that it will maintain its performance given small perturbations to its inputs.  It is this property that has been found to be violated in many instances by adversarial perturbations~\cite{biggio2013evasion,szegedy2013intriguing}.  An adversarial perturbation for a correctly classified input, $x$, is a small perturbation, $r$ satisfying $\|r\|<\epsilon$, that when applied to the input results in an incorrect classification decision, i.e., $f_{\Theta}(x+r)\neq f_{\Theta}(x)$ (see Figure~\ref{fig:perturb}).  A classifier for which adversarial perturbations exist for many of its examples is not considered robust since the model can easily be fooled by small changes to the inputs.

    Once a classifier has been trained, an adversarial perturbation for a given input $x$ can be found by using a gradient-based optimization procedure to search for minimum perturbations to the input that maximize the loss function.  Methods range from quick approximations that take only a single gradient step such as Fast Gradient Sign Method (FGSM)~\cite{goodfellow2014explaining}, to solving the full optimization problem as in~\cite{carlini2016towards}.  While FGSM is not guaranteed to find an adversarial example, solving the full optimization problem is nontrivial, so in~\cite{moosavi2016deepfool} the authors propose an alternative method, called DeepFool, that iteratively projects the input onto the decision boundary of the locally-linearized classification model.  Once adversarial perturbations have been found, they can be used in a robustness metric or to create additional training examples for the classifier.  Training on adversarial examples, referred to as \emph{adversarial learning}, has been shown to increase robustness~\cite{goodfellow2014explaining}.

    In general, there is a trade-off between accuracy and robustness for any given classifier. Optimization techniques used during model training tend to craft highly complex decision boundaries in an attempt to precisely differentiate between each class. Such precision leads to greater accuracy yet introduces some deficiencies as those final, tough to classify data points now lie close to the decision boundary, only needing to be ``nudged" slightly in order to be misclassified. In this manner, a highly accurate model may not be robust. By optimizing for robustness, any decision boundary learned to separate the classes is effectually kept from becoming too precise. While less accurate, a robust model is likely far more reliable in execution.

\section{Automatic Target Recognition with Synthetic Aperture Radar}
\label{sec:sar}

\begin{figure}[ht]
        \centering
        \includegraphics[width=0.35\textwidth]{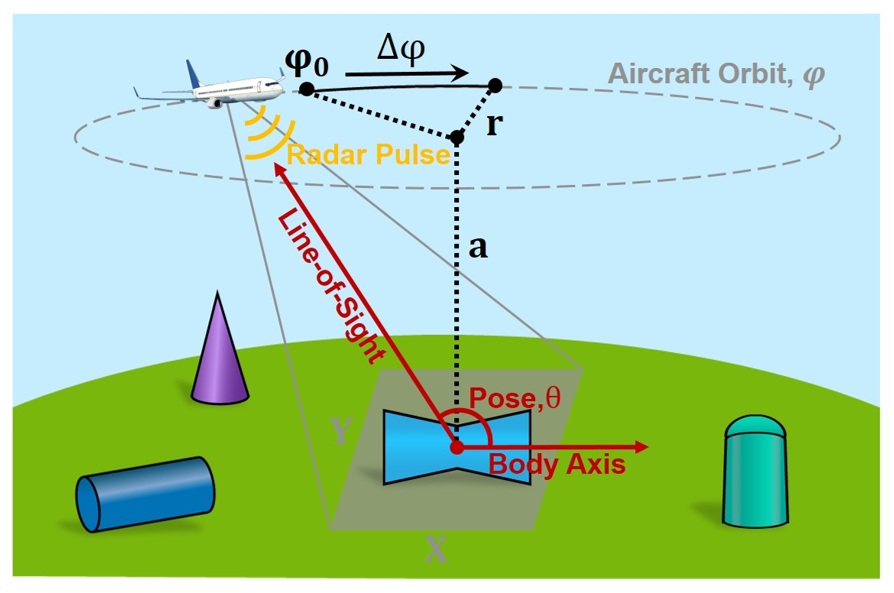}\\
        \hspace{-0.25cm}\includegraphics[width=0.35\textwidth]{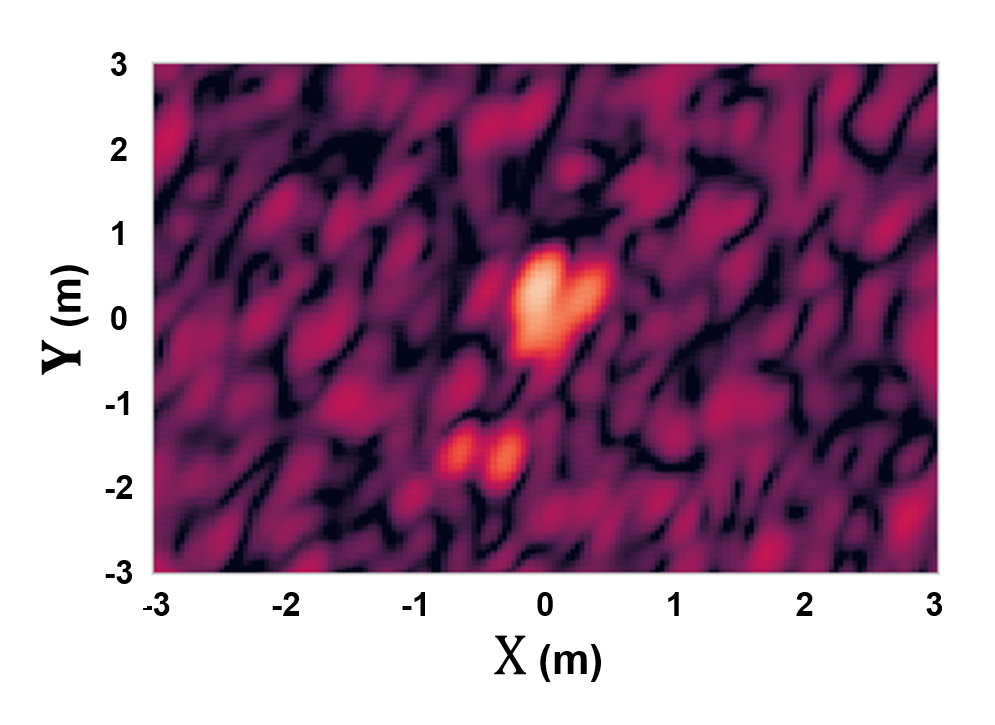}
        \caption{Notional ATR with SAR scenario (top) with simulated SAR image for an hourglass target (bottom)}
        \label{fig:sar_scenario}
    \end{figure}

    We focus on data derived from simulating an airborne SAR, which produces a high-resolution representation of the scene in range and cross-range~\cite{skolnik2008introduction}. Similar to the data set generated in~\cite{dungan2012wide}, we consider the SAR scenario illustrated in Figure~\ref{fig:sar_scenario} with an example input image for an hourglass-shaped target.  The target shapes we consider are described in detail Section~\ref{sec:data_shapes}.  The aircraft flies in a circular orbit around the target of interest while sending Linear Frequency Modulated (LFM) pulses to the target, and collecting the received backscattered pulses.  A SAR image is generated from the complex-valued frequency history of a given orbital segment using back-projection around the target (i.e., spotlight extraction). The goal of the ATR classifier is to determine the target class given the normalized magnitude of the image.

    \subsection{Overview of SAR}

    The principle behind SAR is is to use a traditional mono-static radar with a LFM pulse that provides high range resolution and utilize the motion path of the host platform to produce an ``simulated'' large aperture that can also provide high cross-range resolution.  Without the motion path the angular resolution of the mono-static processed data will be coarse. For a SAR platform following a motion path, and observing a stationary target, the antenna phase center is defined as,
    \begin{equation}
        X_p = [x_p, y_p, z_p]
    \end{equation}
    where there are $p=1,\dots,N_p$ collects across the across the synthetic aperture. The distance to the radar phase center is then given by
    \begin{equation}
        R_p = \sqrt{(x_p - x_0)^2 + (y_p - y_0)^2 + (z_p - z_0)^2}
    \end{equation}
    where the position of the stationary target is $X_0 = [x_0, y_0, z_0]$ defined to be the geometric center of the target shape.  The output of the receiver at time $t_p$ is a sequence of frequency samples delayed by the round trip time between the transmitted signal and the back-scattered response of the target.  There are $K$ frequency samples per received signal denoted by $f_k$. The received signal for each sequence can formulated as
    \begin{equation}
        E_{k, p} = E_T(f_k) \exp{(-i 4 \pi f_k R_p / c)}
    \end{equation}
    where $E_T$ is the complex response for target $T$ and $R_p$ is the distance to the phase center defined above.  There exists a number of techniques to convert these complex valued frequency histories into a 2D image. We utilize the back-projection algorithm described in~\cite{gorham2010sar}.

    \subsection{Target Classification Architecture}
    \label{sec:arch}

    Given a set of complex valued frequency histories across a synthetic aperture illuminating a single target, $S=\{E_{k,p}| k=1,\dots,K \textrm{ and } p=1,\dots,N_p\}$, execute the back-projection algorithm to produce the complex valued image,
    {\small
    \begin{equation}
        \label{eqn:bp}
        x_c = \texttt{backprojection}(S).
    \end{equation}
    }
    A real-valued image is required for the target classification architecture, we apply the following conversion,
    \begin{equation}
        \label{eqn:norm}
        x = (20 \log_{10}(|x_c|) - \mu) / D,
    \end{equation}
    where $\mu$ is the mean value of the magnitude in decibels for the data set and $D$ is set such that most of the values of $x$ fall within the range of $[-1, 1]$.

    The input image, $x$, is assumed to only contain a single target of class $y$ out $C$ different target classes.  To classify the target within the processed image, execute a neural network consisting of a feature extractor, $f:\mathbb{R}^{N \times N} \to \mathbb{R}^M$, and classifier, $c:\mathbb{R}^M \to \mathbb{R}^C$, given by,
    \begin{align}
        h &= f(x; \Theta_f), \\
        \hat{y} &= c(h; \Theta_c),
    \end{align}
    where $\hat{y}$ is the estimated class probability vector. The functions $f$ and $c$ are neural networks whose parameters, $\Theta_f$ and $\Theta_c$, are estimated by minimizing by the following cross-entropy loss using a form of backpropagation,
    {\small
    \begin{equation}
        L_{clf}(x, y; \Theta_f, \Theta_c) = - \sum_{i}^P \hat{y}_{i} \log (y_i).
    \end{equation}
    }

    \subsection{Robustness Techniques}

    \paragraph{Pose Estimation} The first approach we consider to improve robustness and provide better feature learning is to jointly estimate the target class \emph{and} pose, $\theta\in[0,2\pi]$, which is the angle between the target body axis and radar line-of-sight (see Figure \ref{fig:sar_scenario}). Joint training may improve feature learning by forcing the neural network to output features that best represent the information needed to classify a target \emph{and} estimate its pose.

    We discretize the angle into $T$ bins and use the categorical distribution to estimate the pose, $\hat{\theta}$, via a neural network,
    \begin{equation}
        \hat{\theta} = p(h; \Theta_p) \ , \qquad \theta \in \mathbb{R}^T.
    \end{equation}
    The parameters of the pose estimator is trained along with the feature and classification parameters by minimizing an additional loss function,
    {\small
    \begin{equation}
        L_{pose}(x, y; \Theta_p) = - \sum_{i}^T \hat{\theta}_{i} \log (\theta_i)
    \end{equation}
    }

    \paragraph{Similarity Embedding} It is expected that similar inputs will result in feature vectors that are close given a distance metric. Learning a feature space that embeds this property will improve classifier robustness to small changes in target phenomenology (see appendix of \cite{wang2016theoretical}). Similarity depends on the specific application; for ATR using SAR, we define input similarity based on three properties: 1) targets belong to the same class, 2) targets have similar size, \emph{and} 3) the targets have similar pose. To embed similarity into our network, we define a binary similarity label, $s$, ($0$ if inputs are similar, $1$ if not) and consider the contrastive loss between two extracted feature vectors $h_1$ and $h_2$, from separate SAR images $x_1$ and $x_2$,
    {\small
    \begin{equation}
        L_{sim}(h_1, h_2, s; \Theta_f) = (1 - s) \|h_1-h_2\|_2^2 + s \min(1 - \|h_1-h_2\|_2, 0)^2.
    \end{equation}
    }
    This loss is minimized in conjunction with the classification loss.

    \paragraph{Adversarial Learning} Adversarial Learning with FGSM \cite{goodfellow2014explaining} has shown an ability to improve the robustness of a classification model. We utilize FGSM to perturb the complex-valued target frequency history before back-projection, providing a more "``ealistic'' adversarial perturbation of the target response and aiming to improve robustness against small variations in a target's complex-valued phenomenology. That is, for a radar, we want to consider perturbations of the signal being received at the sensor rather than on the ``pixels'' of the input image to the target classifier. To do this, define the signal processing function, $x = g(s)$, that takes the received signal, $s$, and processes the signal through the back-projection algorithm and normalizing functions defined in Equations~\ref{eqn:bp} and \ref{eqn:norm}. The perturbation defined by FGSM is given by,
    \begin{equation}
        \eta = \epsilon \text{sign}(\nabla_x L_{clf}(x=g(s), y; \Theta_f, \Theta_c)).
    \end{equation}
    Then the perturbed input image the neural network is
    \begin{equation}
        x_{\text{FGSM}}' = g(s + \eta).
    \end{equation}

    For adversarial training, we minimize the additional loss term,
    \begin{equation}
        L_{adv}(x_{\text{FGSM}}', y; \Theta_f, \Theta_c) = - \sum_{i}^P \hat{y'}_{i} \log (y_i)
    \end{equation}
    where $y$ is the label of the original image, $x$, and $\hat{y'}_{i}$ is the label of the perturbed image, $x_{\text{FGSM}}'$.

\section{Experiments}

  \subsection{Data}

  The data set consists of 715 unique 3D targets of various size corresponding to four shape classes shown in Figure~\ref{fig:shapes}: cylinder, cone, dome-cylinder, and hour-glass.  For each individual target, 1000 SAR images are generated, resulting in a total of 715,000 images\footnote{This data set is being prepared for public release with accompanying technical details on how the images are generated and processed as well as how one can go about exploiting the physical structure of the radar signals} to train and test our classifiers. This data set varies the target pose $\theta$, radar altitude $a$, orbital radius $r$, initial orbit location $\varphi_0$, and background noise as shown in Figure~\ref{fig:sar_scenario}. Each SAR image represents a six meter window in the xy-plane with $160$ samples (depiction of SAR scenario and sample image shown in Figure~\ref{fig:sar_scenario}). The normalizing parameters in Equation~\ref{eqn:norm} are $\mu=-40$ and $D=50$.

    \subsection{RF Simulation}
    \label{sec:data_shapes}
    \begin{figure}
        \centering
        \includegraphics[width=0.5\textwidth]{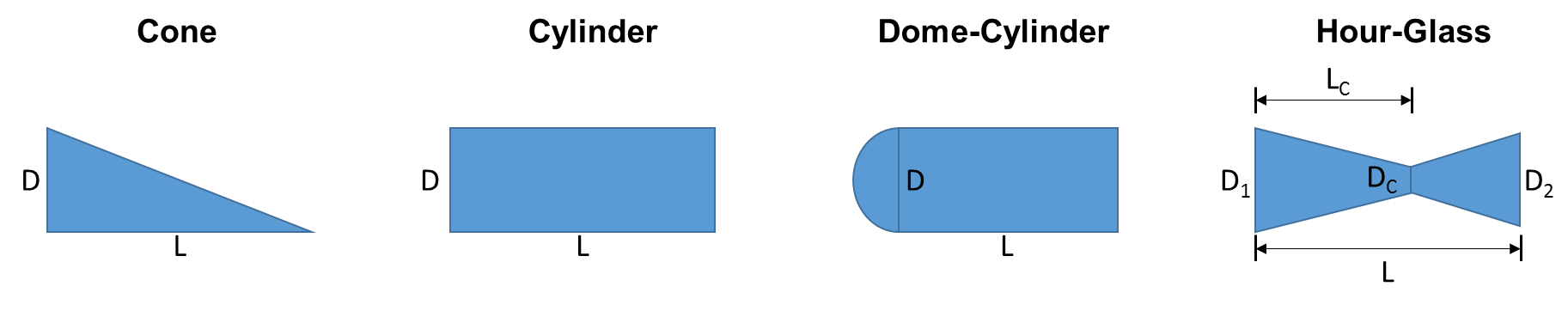}
        \caption{Description of shapes in data set.}
        \label{fig:shapes}
    \end{figure}

    This data set utilizes four different shape classes that are depicted in Figure~\ref{fig:shapes}. These shapes are modeled in 3D by assuming the targets are symmetric along the body-axis (roll symmetric). To generate a random sample of target shapes we define the distribution of parameters shown in Figure~\ref{fig:shapes} as:
    {\small
    \begin{align}
        L &\sim U[1,4] \\
        D &\sim U[1, 2] \\
        D_1 &\sim U[1, 2] \\
        D_2 &\sim U[1, 2] \\
        L_C &= L/2 + 0.1 \\
        D_C & =0.1.
    \end{align}
    }

    The distribution of parameters are defined to challenge a classification algorithm to estimate the shape class independent of the objects sizes.  In addition to modeling the geometric shape, half the samples will include basic ``ring'' (e.g., notch or groove) randomly along the body axis and is also roll symmetric.

\begin{figure}[ht]
        \centering
        \includegraphics[width=0.35\linewidth]{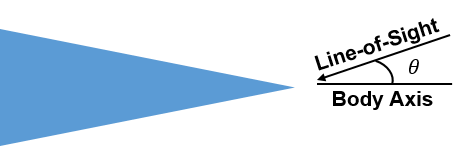}
        \caption{Description of line-of-sight.}
        \label{fig:los}
    \end{figure}

    We utilize an RF simulation tool developed internally to provide the frequency response for a given look angle and target shape.  The simulations utilize Geometric Diffraction Theory (GDT) \cite{keller1962Geometrical} to model the responses for a select number of scattering centers for a given shape.  GDT is applicable in the high frequency region we focus on in this data set, the electromagnetic field can be written as
    \begin{equation}
        E(f_k, s_n) = A(f_k, s_n, \theta) \exp{(-i 4 \pi f_k r_n / c)}
    \end{equation}
    where there are $K$ frequency samples per signal denoted by $f_k$, $s_n$ is the scattering center at a given range $r_n$, $A(f_k, s_n, \theta)$ is the complex amplitude response of the scattering center for given line-of-sight $\theta$ to the radar (see Figure~\ref{fig:los}), and $c$ is the speed of light constant. See \cite{bechtel1965application} and \cite{senoir1973further} for an example of how to model the complex amplitude of cones.

    \begin{table*}[t]
    \begin{small}
    \begin{center}
    \caption{Summary of accuracy and robustness results of augmented classifiers for ATR with SAR}
    \label{tab:exp_results}
    \begin{tabular}{l|c|cr}
    &{Accuracy} & & {Robustness} \\ \hline
    BASIC & 0.896 $\pm$ 0.011 & & 0.0201 $\pm$ 0.0011 \\
    POSE & 0.899 $\pm$ 0.009 & & 0.0209 $\pm$ 0.0006 \\
    SIM & 0.921 $\pm$ 0.013 & & 0.0204 $\pm$ 0.0008 \\
    POSE+SIM & 0.912 $\pm$ 0.013 & & 0.0204 $\pm$ 0.0018  \\
    ADV & 0.871 $\pm$ 0.006 & & 0.0213 $\pm$ 0.0011 \\
    ADV+SIM & 0.889 $\pm$ 0.005 & & 0.0224 $\pm$ 0.0026 \\
    \end{tabular}
    \end{center}
    \end{small}
    \end{table*}

    An RF simulation is conducted for a randomly sampled shape, frequencies, and rotation angles about the geometric center of the object. For this data set, the center frequency is $24 \thinspace  \mathrm{GHz}$, bandwidth is $B=0.5 \thinspace \mathrm{GHz}$, and the number of frequency samples is $K=64$. A simulation for a sampled target, $T$, is then given by
    \begin{equation}
        E_T^P(f_k, \phi) = \sum_{n=1}^{N} A(f_k, s_n, \theta) \exp{(-i 4 \pi f_k r_n / c)}
    \end{equation}
    where the scattering center range, $r_n$, is defined to be relative to the geometric center of the target, and $P$ is one of the four possible polarization combinations: HH, HV, VH, and VV. For images generated in this paper, we utilize the circular polarized signal: $E^T = 0.5 (E_T^{HH} + E_T^{VV})$.

  \subsection{Target Classification Model}
    The feature extraction architecture is a simple convolutional neural network (CNN) with layers C(16, 20, 1, 0) - C(32, 3, 2, 1) - C(64, 3, 2, 1) - C(128, 3, 2, 1) - C(256, 3, 2, 1) - P(5), where C(n, k, s, p) is a convolution layer followed by a ReLU non-linearity where $n$ is the number of output channels, $k$ is the kernels size in both dimensions, $s$ is the stride, and $p$ is the padding.  The last layer is an average pooling layer with a kernel size such that the output is a vector of size $256$.

    The classifier function is a fully connected neural network of two linear layers: L(64) followed by a ReLU, and L(4) followed by a soft-max layer.

    The pose estimator is a fully connected neural network of two linear layers: L(64) followed by a ReLU, and L(180) followed by a soft-max. We discretize the angle space into $T=180$ angle bins.

  \subsection{Evaluation Metrics}

  We perform 4-fold cross-validation to train and estimate the out-of-sample accuracy of each classifier. To evaluate the robustness of a classifier to adversarial perturbations, we use the metric, $\hat{\rho}_{adv}(f)$, introduced in ~\cite{moosavi2016deepfool}:

  {\small
  \begin{equation}
  \hat{\rho}_{adv}(f) = \frac{1}{|D|} \sum_{x\epsilon D}\frac{\|\hat{r}(x)\|_{2}}{\|x\|_{2}}.
  \end{equation}
  }
  $\|\cdot\|_2$ represents the Euclidean (i.e., L2) norm.  The minimum adversarial perturbation, $\hat{r}(x)$, for each SAR image, $x$, in the validation data set, $D$, is computed using the DeepFool algorithm. When comparing two classifiers, if $\hat{\rho}_{adv}(f_1) > \hat{\rho}_{adv}(f_2)$, we conclude classifier $f_1$ is more robust than $f_2$.

  \subsection{Results}

     \begin{figure}[ht]
      \centering
      \includegraphics[width=0.7\linewidth]{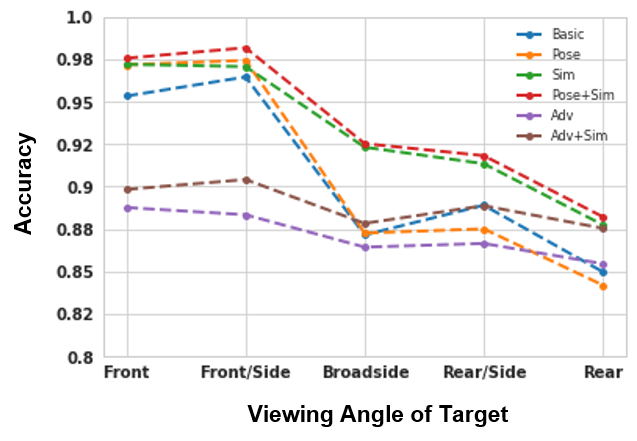} \\
      \hspace{-0.25cm}\includegraphics[width=0.7\linewidth]{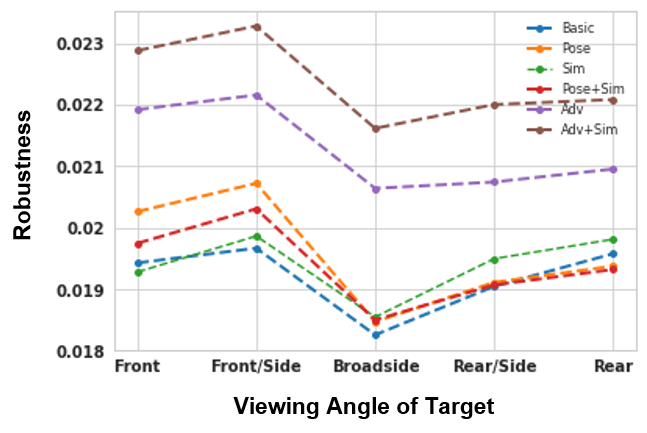}
      \caption{Accuracy (top) and Robustness (bottom) results of the described classification architectures for ATR of the simulated SAR images}
      \label{fig:results}
    \end{figure}

   We compare the basic architecture of feature extractor followed by classifier (BASIC) with the following augmented training schemes: pose estimation (POSE), similarity embedding (SIM), pose estimation and similarity embedding (POSE+SIM), adversarial learning with FGSM (ADV), and adversarial learning and similarity embedding (ADV+SIM). Overall results are shown in Table~\ref{tab:exp_results}.

   Each of the augmented training techniques leads to an increase in robustness over the basic classifier. Adversarial learning with FGSM with similarity embedding has the highest robustness to adversarial perturbations, which is expected because this approach directly optimizes a loss function that applies small perturbations to the classifier input.  However, since FGSM is a form of regularization, adversarial learning results in a slight drop in accuracy compared to the basic classifier. On the other hand, pose estimation and similarity embedding both result in increases in accuracy. We theorize that by conditioning the classifier on information such as pose and similarity properties, we learn more effective representations of the data and hence achieve higher accuracy.

    Since each image is a randomly sampled target with a random pose (or look angle), results can be broken down to performance across this viewing angle (see Figure~\ref{fig:los}).  Figure~\ref{fig:results} illustrates the varying performance of the classifiers binned across five different viewing geometries. Front and front/side viewing angles provide the highest amount variability between each of the four object shapes and therefore we expect higher accuracy over other viewing geometries.  Since all shapes look similar in the rear viewing geometry, e.g., the base of the cone and cylinder, we expect a drop in accuracy compared to front viewing. For robustness, there is a drop in performance around broadside geometries.  Broadside is usually specular in nature and therefore exhibits little phenomenology to classify shape, therefore small variations in the signal can lead to miss-classification.  Examining the behavior across the different training methods, adversarial learning clearly improves the robustness of the classifier over all viewing angles while reducing the overall accuracy.  Yet, the results demonstrate that adding similarity embedding to adversarial learning improves the robustness while also improving the accuracy over basic adversarial learning. Adding pose is expected to improve accuracy and remains to be seen in future analyses if robustness also improves.

\section{Discussion and Future Work}

    In this work, we present a convolutional neural network architecture and selection of training techniques for learning accurate and robust representations of 3D targets in active sensing environments, such as radar.  We investigate these techniques using a simulated SAR for ATR scenario, and find adversarial learning to be the approach that achieves the highest robustness to adversarial attack, while pose estimation with similarity embedding increases the robustness while also achieving the highest accuracy.  Future work will include incorporating additional robustness metrics, performing similar analysis on other existing radar data sets (e.g., MSTAR~\cite{yang2005automatic}), and exploring the applicability of generative modeling for adversarial data augmentation that avoid the need to calculate the gradient. In-line with other current work, we are developing generative models, such as Generative Adversarial Networks (GAN)~\cite{goodfellow2014generative}, to sample simulated radar observations that are within the target distribution but fool our classifiers~\cite{guo2017synthetic, xiao2018generating}.  We hypothesize that training with these generative models will help increase robustness in data-starved radar applications.

\bibliographystyle{IEEEtran}
\bibliography{bibliography}

\end{document}